\documentclass[journal]{IEEEtran}
\usepackage{graphics} 
\usepackage{epsfig} 
\usepackage{amsmath} 
\usepackage{amssymb}  
\usepackage{mathtools}
\usepackage{colortbl}
\usepackage{tabularx}
\usepackage{multirow}
\usepackage{url}

\newcommand\kevinupdate[1]{\textcolor{black}{#1}}

\newcommand{\insteadof}[1]{\ignorespaces}

\ifCLASSINFOpdf
\else
\fi
\begin{document}
%
\title{Recurrent-OctoMap: Learning State-based Map Refinement for Long-Term Semantic Mapping with 3D-Lidar Data}
%
%
%

\author{Li~Sun$^{1}$, 
             Zhi~Yan$^{2}$,
             Anestis~Zaganidis$^{1}$,
			Cheng~Zhao$^{1}$, and
			Tom~Duckett$^{1}$
\thanks{Manuscript received: February 24, 2018; Revised April 23, 2018; Accepted June 19, 2018.}
\thanks{This paper was recommended for publication by Editor Dongheui Lee upon evaluation of the Associate Editor and Reviwers’comments.}
\thanks{$^{1}$L. Sun, C. Cheng, A. Zaganidis, T. Duckett was with Lincoln Centre for Autonomous Systems (L-CAS), University of Lincoln, UK email: lsun, azaganidis, czhao, tduckett@lincoln.ac.uk}
\thanks{$^{2}$Z. Yan was with LE2I-CNRS, University of Technology of Belfort-Montb\'eliard (UTBM), France email: zhi.yan@utbm.fr}%
\thanks{Digital Object Identifier (DOI): see top of this page.}
}

%
%

\markboth{IEEE Robotics and Automation Letters. Preprint Version. Accepted June, 2018}%
{Shell \MakeLowercase{\textit{et al.}}: Bare Demo of IEEEtran.cls for IEEE Journals}
%



\maketitle

\begin{abstract}
This paper presents a novel semantic mapping approach, Recurrent-OctoMap, learned from long-term 3D Lidar data. Most existing semantic mapping approaches focus on improving semantic understanding of single frames, rather than 3D refinement of semantic maps (i.e. fusing semantic observations). The most widely-used approach for 3D semantic map refinement is a Bayesian update, which fuses the consecutive predictive probabilities following a Markov-Chain model. 
Instead, we propose a learning approach to fuse the semantic features, rather than simply fusing predictions from a classifier. 
In our approach, we represent and maintain our 3D map as an OctoMap, and model each cell as a recurrent neural network (RNN), to obtain a Recurrent-OctoMap. In this case, the semantic mapping process can be formulated as a sequence-to-sequence encoding-decoding problem. Moreover, in order to extend the duration of observations in our Recurrent-OctoMap, we developed a robust 3D localization and mapping system for successively mapping a dynamic environment using more than two weeks of data, and the system can be trained and deployed with arbitrary memory length.
We validate our approach on the ETH long-term 3D Lidar dataset~\cite{eth-data}. The experimental results show that our proposed approach outperforms the conventional ``Bayesian update" approach.

\end{abstract}


%
\IEEEpeerreviewmaketitle

\section{INTRODUCTION}
\label{sec:introduction}

\IEEEPARstart{C}{ompared} to the more mature research on semantic scene understanding and Simultaneous Localization and Mapping (SLAM), robust semantic mapping is still an open problem. While a conventional 3D map is useful for robot localization and navigation, a 3D semantic map has great potential to further improve the robustness of localization in changing environments and help the robot to consider semantics and dynamics in task and motion planning. Most existing research on semantic mapping considers indoor scenes, while there are few approaches for large-scale outdoor environments. For indoor semantic mapping, methods such as RGBD-SLAM or Kinect-Fusion are widely used, while research on outdoor semantic mapping employs stereo-based mapping or 3D-Lidar-based mapping. In the existing semantic mapping approaches, 2D RGB-based semantic segmentation methods, e.g. Fully Convolutional Neural Networks, are typically adapted. The 2D semantic label can be transferred into 3D through visual geometry. There are a few approaches working on 3D refinement for semantic mapping, where Markov-Chain-based Bayesian updates~\cite{bayes-update} were used for fusion of consecutive semantic labels in~\cite{dense2014icra}, and then widely-used in most of the following research \cite{oxford3d,dense2014icra,ic2017icra,zhao2017fully,xiangyu2017rnn,ma2017iros}. 

\begin{figure}[t]
\centering
\includegraphics[width=.49\textwidth]{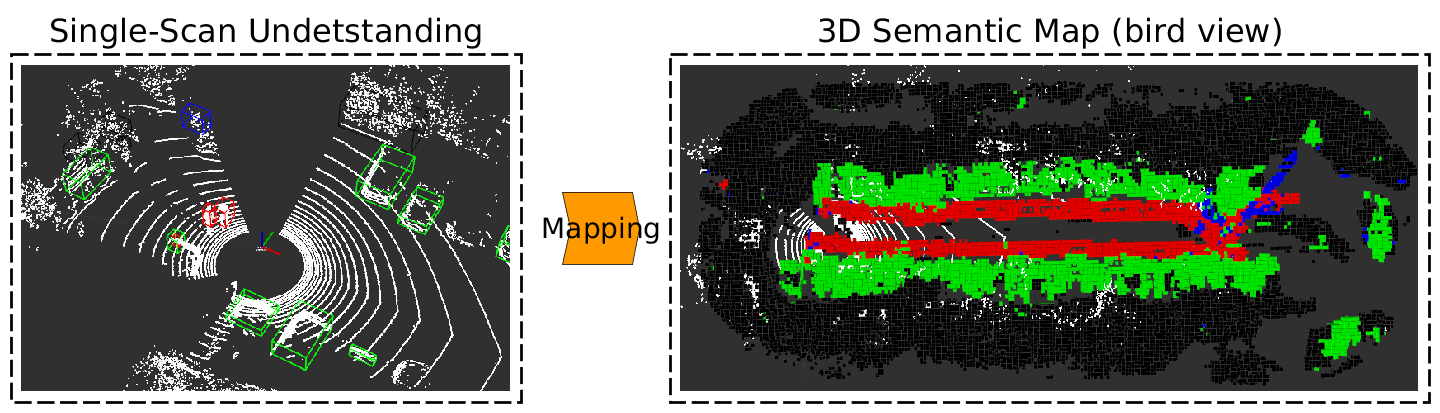}
\caption{Overview of long-term semantic mapping with 3D Lidar data.}
\label{fig:overview}
\end{figure}

Pioneering research~\cite{xiangyu2017rnn,ma2017iros} uses visual odometry to improve 2D semantic segmentation in RGB-D videos. Unfortunately, because of the limited field of view (FOV) of RGB-D cameras, limited pixels can be associated in a long sequence. Hence, the improvement obtained from fusing consecutive data in their research is not very substantial. 3D Lidar sensors such as Velodyne have $360^{\circ}$ FOV, which combined with precise odometry made long-term mapping possible~\cite{eth-data}. Moreover, with the development of geometry-based semantic understanding~\cite{pntnet}, semantic mapping has the potential to be achieved using a single 3D Lidar sensor.

Most of the existing approaches have the following limitations: Firstly, both indoor and the outdoor semantic mapping approaches generally rely on RGB-based scene understanding. Secondly, they simply fuse the predicted probabilities from the classifier with a Bayesian update for the refinement of the 3D semantic map. Thirdly, only short-term consecutive observations are used for semantic fusion, while the long-term spatio-temporal semantic correlation, e.g. from consecutive minutes, hours and days, is neglected. 

In this research, we use a single 3D Lidar for semantic scene understanding, localization and mapping (Fig. \ref{fig:overview}). A long-term dataset (up to two weeks) is used to learn the spatio-temporal semantics. The main contributions of this paper are two-fold: Firstly, we propose a novel approach, Recurrent-OctoMap, to fuse the semantic observations. Our semantic map is state-based, maintainable, and with flexible memory duration. Secondly, our semantic mapping approach uses a single 3D-Lidar and no RGB camera is required.


\section{RELATED WORK}
\label{sec:related_work}

\subsection{3D Semantic Mapping}
The first 3D indoor dense semantic mapping was proposed in \cite{dense2014icra}. In this research, a Bayesian update~\cite{bayes-update} is first adapted to 3D map refinement and a 3D Conditional Random Field (CRF) is further used to optimize the semantic predictions for adjacent voxels.

Tateno et al.~\cite{tateno2016japanese} proposed an approach to map the detected objects into a dense 3D map. In their approach, 3D-based template matching is used to register the scene objects with pre-scanned models (known objects). 
McCormac et al.~\cite{ic2017icra} first adapted a deep learning-based segmentation approach to 3D semantic mapping in indoor scenes. Cheng et al.~\cite{zhao2017fully} first proposed a 3D semantic mapping pipeline for material understanding of an indoor scene. In their approach, a more advanced  neural network with boundary-optimization is used for semantic segmentation.

Pioneering researchers \cite{xiangyu2017rnn,ma2017iros} investigated improving the 2D scene understanding using consecutive frames of observations. In their research, RGBD-based visual odometry is used to associate the pixels between consecutive frames and a RNN is trained from multiple observations for semantic classification of the latest frame.

Compared to the work on indoor 3D semantic mapping, there are fewer approaches for outdoor 3D semantic mapping~\cite{oxford2d,oxford3d}. In \cite{oxford2d}, the street semantics are obtained by 2D semantic segmentation from a RGB camera on a driving vehicle, and a dense 2D ground-plane semantic map can be obtained from multi-view imagery. They further extend the 2D on-road semantic mapping to dense 3D mapping \cite{oxford3d}, where stereo images are used for dense 3D reconstruction. In \cite{outdoor2018semantic}, a semantic map can be built from multi-sensory data for navigation in an off-road environment, where the scene understanding is obtained incrementally from 2D RGB images and 3D Lidar is employed for dense 3D mapping. 

\subsection{3D Lidar-based Object Detection and Scene Understanding}
Model-free segmentation methods (i.e. clustering-based) are widely used for objectness detection in 3D Lidar data \cite{depth-clustering,yan2017online}. 
Bogoslavskyi et.al~\cite{depth-clustering} developed a fast method with small computational demands through converting 3D Lidar scans into 2D range images. Yan et al.~\cite{yan2017online} proposed an adaptive clustering approach that enables to use different optimal thresholds for point cloud clustering according to the scan ranges.
Moreover, Dewan et al.~\cite{dewan2016motion} proposed a model-free approach for detecting and tracking dynamic objects, which relies only on motion cues.
The conventional approach for 3D Lidar-based object recognition employs hand-crafted features~\cite{navarro-serment09fsr,kidono11iv,spinello11icra,deuge13acra}, such as PCA, 3D grid features, hierarchical part features, etc.


Most of the model-based detection approaches convert the 3D Lidar point cloud into a 2D image~\cite{chen2017birdview,squeeze-seg} or 3D voxel grid~\cite{3dcnn}, in order to employ a CNN-based method to learn pixel-wise semantic segmentation or multi-box object detection. 
Researchers also started to develop new learning approaches~\cite{pntnet} that are applied directly to the irregular 3D point cloud. In PointNet~\cite{pntnet}, a fully connected neural network can be learned from geometry-only data for semantic classification, and a max-pooling function is used to achieve order invariance. Engelmann et al.~\cite{context2017exploring} further explore the spatial context among different semantic categories using a recurrent neural network. The geometry-based learning methods, e.g. \cite{pntnet,context2017exploring}, can learn the semantics by exploiting the inherent 3D geometry structure of the point cloud data. Therefore, these methods have the potential to be adapted from Kinect-type point clouds to 3D Lidar point clouds.

\subsection{Long-term Mapping and Persistent Mapping}

Over the past two decades of development of Simultaneous Localization and Mapping (SLAM), 
approaches such as 2D laser-based SLAM~\cite{grisetti2007gmapping}, stereo-based visual-SLAM~\cite{mur2017orb} and 3D Lidar SLAM~\cite{zhang2014loam} 
have been reaching the maturity required for industrial applications.
Some interesting problems also occur in long-term mapping of dynamic environments: 
e.g.\
long-term maintenance of the map in a changing environment~\cite{eth-data};
improvement of the robot's odometry and removing the effects of dynamic objects~\cite{einhorn2015generic}; and identification of the environment dynamics~\cite{fremen}.

\begin{figure*}[thpb]
\centering
\includegraphics[width=1.0\textwidth]{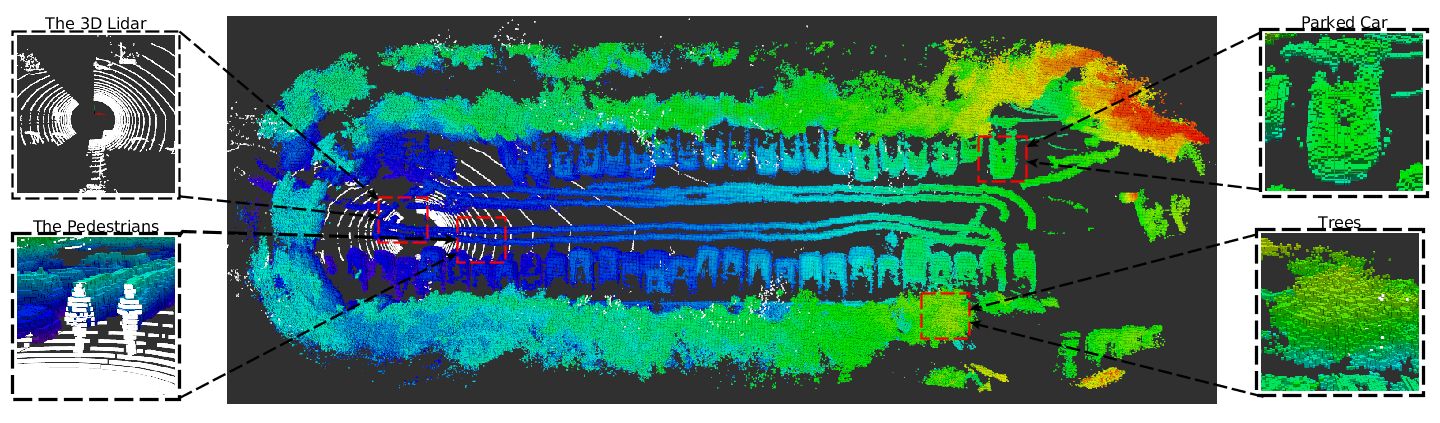}
\caption{An example of Parking Lot dataset. The map of day 1 is shown. In this dataset, the Lidar sensor only provide geometry readings (i.e. X, Y, Z), and both RGB and intensity are not available. }
\label{fig:parking-lot}
\end{figure*}

\subsection{Discussion}
Bayesian update and CRF-based approaches are used for 3D map refinement in almost all previous semantic mapping research. The CRF-based approach is a pair-wise optimization mainly for object boundary refinement in image-based semantic segmentation. Bayesian update is the only proposed approach for probabilistic fusion of semantic observations in 3D semantic mapping. Researchers have started using RNN-based models to improve single-frame segmentation performance via associating consecutive frames in RGB-D video. However, because of the constraints of the limited field of view and odometry precision, these methods are not practical in large-scale outdoor 3D semantic mapping. 

Both the existing indoor and outdoor 3D semantic mapping approaches rely on RGB-based semantic understanding. With the development of 3D Lidar-based odometry~\cite{zhang2014loam} and geometry-based semantic understanding, outdoor semantic mapping thus has the potential to be accomplished using a single 3D Lidar. Moreover, in the existing semantic map research, only short-term semantic fusion is considered. Through long-term mapping, the spatial-temporal semantics obtained over longer time scales can be tracked and exploited. This paper thus aims to achieve 3D semantic mapping using a single 3D Lidar and proposes a novel approach for state-based fusion learned from long-term data.

\section{METHODOLOGY}
\label{sec:methodology}


\subsection{Problem Formulation}\label{sec:problem-formulation}
Given a sequence of observations $o_0^t = [o_0, ..., o_t]$ of a voxel in the 3D map, the goal of semantic fusion is to obtain the predicted probabilities  $p(c_t|o_0^t)$ for all the semantic classes $c_t$ depending on this sequence of observations $o_0^t$. The conventional Bayesian update fuses the predicted probabilities through a first order Markov assumption \cite{eth-data}:

\begin{equation}
P(c_t|o_0^t) = \dfrac{1}{Z'} P(c_t|o_t)P(c_{t-1}|o_0^{t-1})
\label{eq:Bayesian update}
\end{equation}  

\noindent where, $Z'$ is a normalization term. Instead of end-prediction fusion, we use a high-dimensional hidden state $state_t$ to assist the semantic fusion:

\begin{equation}
\begin{split}
P(c_t|o_0^t) = Logit\big(Decoder(state_t)\big) \\
state_{t} \leftarrow RNN(state_{t-1}, o_t)~~~~~~~~\\
\end{split}
\end{equation}  

\noindent where we use a Decoder to approximate the prediction, and a Logistic function is used to squash the activation values to normalized predicted probabilities. We model the semantic fusion as a transition of the hidden state from $state_{t-1}$ to $state_t$. In our approach, a RNN (Recurrent Neural Network) is used as the transition function and the semantic observation $o_t$ provides the input to the RNN.

\subsection{Single-Scan Semantic Understanding}\label{sec:single-scan-understanding}\label{sec:semantic_understanding}

\begin{figure*}[thpb]
\centering
\includegraphics[width=.9\textwidth]{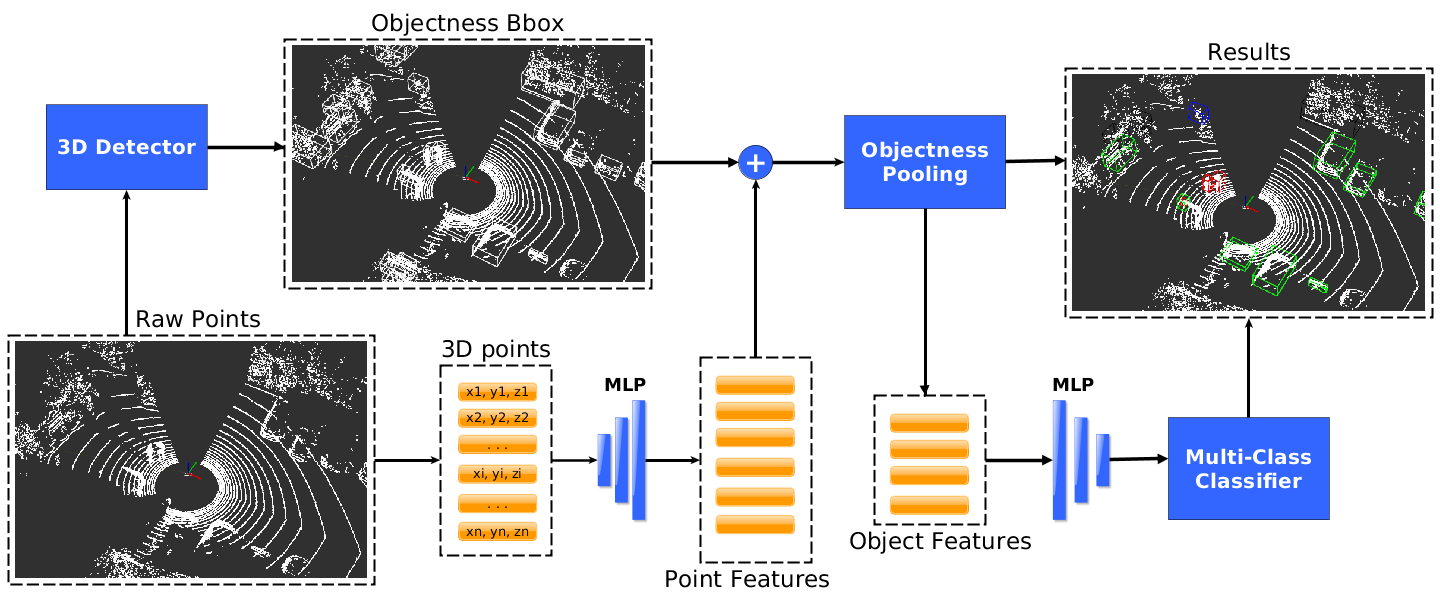}
\caption{The pipeline of the single-scan semantic understanding (object detection). The first MLP uses a structure of five layers, with hidden layers of size 64, 64, 64, 128 and 1024. The second MLP is a two-layer structure and the hidden layer sizes are 512 and 256. $relu$ is used as the activation function.}
\label{fig:detection_pipeline}
\end{figure*}

\subsubsection{Object Detection}
An overview of the proposed detection pipeline is shown in Fig.~\ref{fig:detection_pipeline}.
In our approach, we incorporate model-free objectness detection and a fully connected network as an object detection pipeline. A model-free 3D approach~\cite{yan2017online} is employed for objectness detection. We adapt Point-Net~\cite{pntnet} for object recognition in a single 3D Lidar scan. A multi-layer perceptron (MLP) is connected to all the 3D points of a Lidar scan to learn the non-linear feature embeddings of the point-level features. We apply max-pooling within the 3D bounding boxes obtained from the detector and another multi-layer perceptron is used to learn object-level features. The network output is finally connected with the ground truth labels using a multi-class softmax loss function. To be more specific, given a set of 3D points $[p_i^x, p_i^y, p_i^z]_{N_p}$ in a Lidar scan, the point features $[f_i^{p}]_{N_p}$ can be obtained via the first MLP $mlp_p$:
\begin{equation}
[f_i^{p}]_{N_p} = mlp_p([p_i^x, p_i^y, p_i^z]_{N_p})
\end{equation}
\noindent With the objectness bounding box obtained from the 3D detector, we apply objectness-pooling (i.e. max-pooling within the objectness bounding box) and the object features $[f_j^{o}]_{N_o}$ can be obtained by the second MLP $mlp_o$:
\begin{equation}
[f_j^{o}]_{N_o} = mlp_o\big(pooling_{obj}([f_p])\big)
\end{equation}
\noindent Finally, the negative log likelihood of all the labeled semantic objects are minimized for the whole dataset: 
\begin{equation}\label{eq:loss1}
loss =-\sum_{k}^{N_c} log \mathcal{L} \big(softmax([f_j^{o}]_{N_o}), [label_j]_{N_o}\big)
\end{equation}
\noindent $N_c$ is the number of Lidar point clouds (scans). We propagate the object semantic feature $f_j^{o}$ to all the points within the bounding box. Thereby, each point $<x_i, y_i, z_i>$ of a Lidar scan will have a semantic feature $f_{<x_i, y_i, z_i>}$.

\subsubsection{Transfer Learning from KITTI Dataset}
We pre-train our detection network on the KITTI dataset\footnote{http://www.cvlibs.net/datasets/kitti/}, as large-scale manual annotated examples are available. The 3D Lidar sensor used in KITTI is a Velodyne VLP64E, while our application uses a Velodyne HDL32. Since they have different resolutions and fields of view\footnote{VLP64E: -24.9-2 vertical FOV; HDL32: -30.67-10.67 vertical FOV.}, we convert the KITTI point cloud into rings and down-sample the rings depending on the HDL32's vertical angle interval in order to eliminate the data differences. Moreover, we include random rotation along the z-axis of the world frame as KITTI only has front-view annotations. Having trained the neural network on the KITTI data, we further fine-tune the model using our annotations of ETH parking-lot dataset~\cite{eth-data}.

We chose the combination of 3D clustering for objectness detection and learning of features for recognition in order to maximize the performance with limited data annotations. We also tried an end-to-end approach, e.g. \cite{chen2017birdview,3dcnn}, but unfortunately these approaches cannot produce satisfactory results in our application because of their sensitivity to the sensor configuration. It is worth noting that our proposed Recurrent-OctoMap is not constrained to specific types of single-scan semantic understanding approaches. Either object detection or semantic segmentation approaches can be employed to produce the semantic observations (features) as the input to Recurrent-OctoMap.

\begin{figure}[thpb]
\centering
\includegraphics[width=0.46\textwidth]{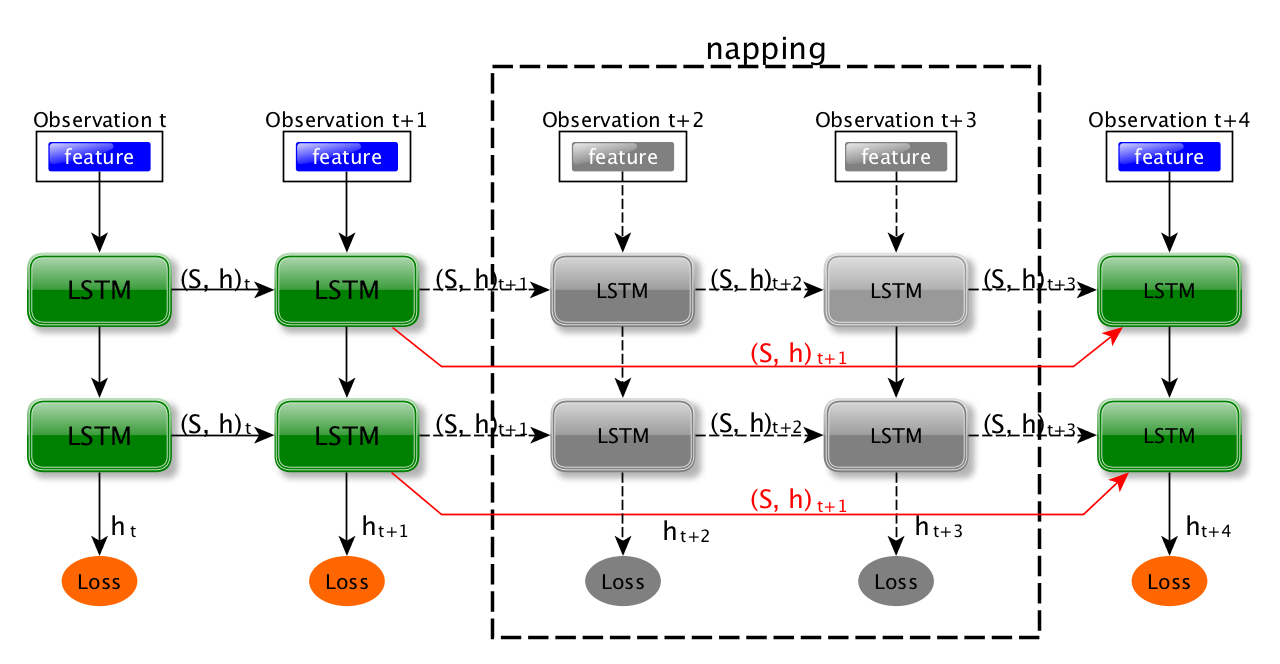}
\caption{The architecture of our Nap-LSTM. This figure shows an example where the state of $t_1$ will pass to $t_4$ when the observations at $t_2$ and $t_3$ are not available. In this case, MNTD must be at least 2, otherwise the state of $t_4$ will be initialized to zeros.}
\label{fig:nap-lstm}
\end{figure}

\subsection{Long-term 3D Lidar Localization and Mapping}\label{sec:long-term-slam}\label{sec:long-term-mapping}

\subsubsection{3D-Lidar-based Localization}
LOAM~\cite{zhang2014loam} provides a robust 2D Laser/3D Lidar odometry using state-based Iterative Close Point (ICP) registration. In this research, edge points and plane points are extracted for registration. The motion estimation in LOAM has two steps: registering the current scan to the previous one as an initial guess, and registering the current scan to the map.

\subsubsection{Long-term 3D Mapping and Map Maintenance}\label{sec:lmmm}
It is really important to maintain the 3D map in the long term. In this paper, our Recurrent-OctoMap inherits the occupancy functions from OctoMap~\cite{octomap} to regularize the voxel-wise semantic observations, store the semantic states, and detect the dynamic changes. Given a new Lidar scan \mbox{$[<x_i, y_i, z_i>]_{N_p}$} and the Lidar odometry obtained from LOAM $T_t$, we apply an inverse transform to transform the point cloud to the map frame \mbox{$[<x_i', y_i', z_i'>]_{N_p}$}, and assign to each Recurrent-OctoMap cell a semantic feature:
\begin{equation}
f_{cell} = ave\_pooling([f_{(<x_i', y_i', z_i'>)}]_{N_{cell}})
\label{eq:cell_feat}
\end{equation}
\noindent where $N_{cell}$ is the number of points in a cell and $ave\_pooling$ is the average pooling function.

As there are dynamic objects, e.g. pedestrians, cyclists and driving cars, for visualization purposes we introduce a minimum time period for objects to remain in the map (approximately five minutes in our experiments).


\subsection{Semantic Mapping based on Recurrent-OctoMap}\label{sec:recurrent-octomap}

\begin{figure*}[thpb]
\centering
\includegraphics[width=0.85\textwidth]{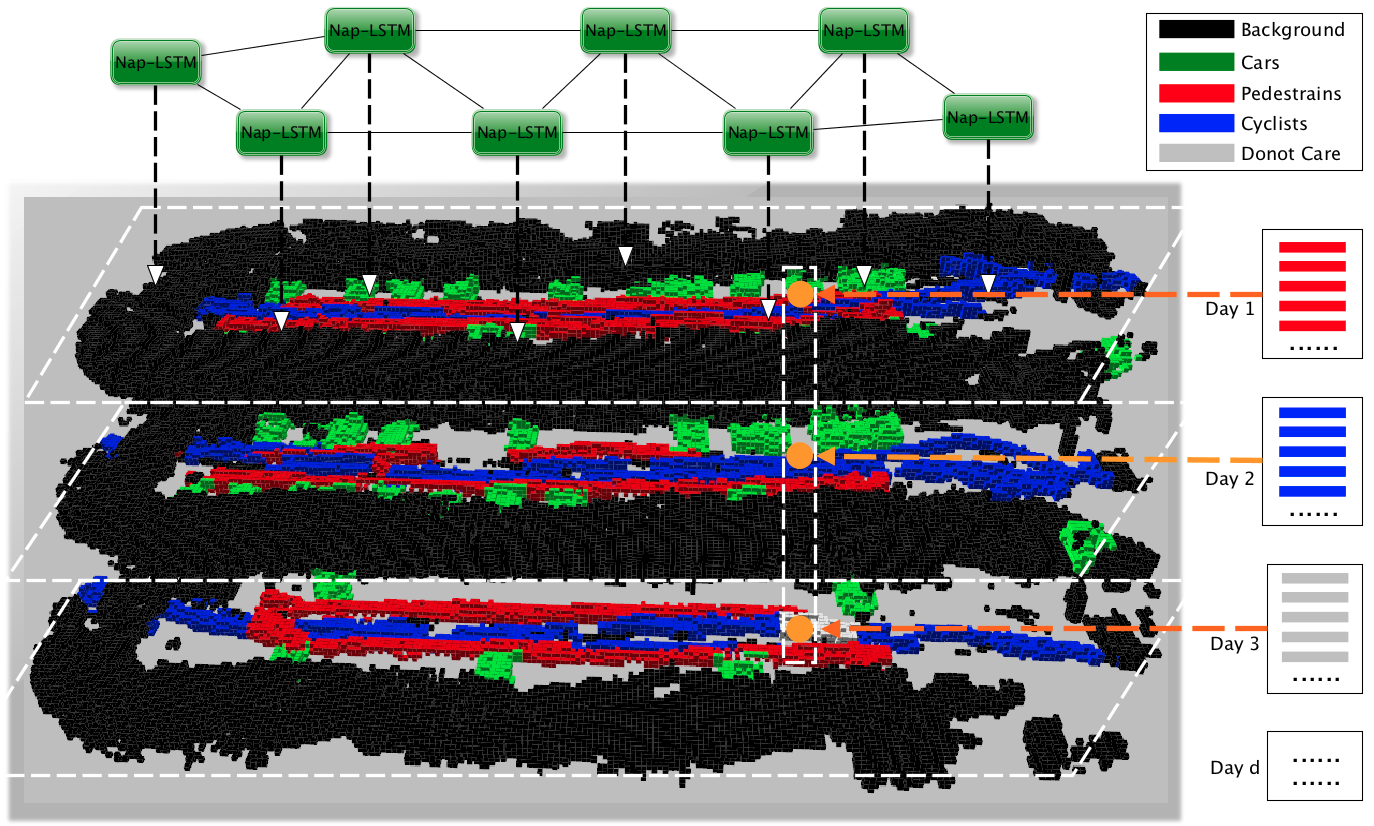}
\caption{Learning Recurrent-OctoMap from long-term semantic mapping. For some locations the semantic categories are constant, while some are changing during different days (take the site in orange for an example). \kevinupdate{All of the LSTMs share the same weights, as indicated by the solid lines.}}
\label{fig:long-term-rnn}
\end{figure*}

\subsubsection{Semantic Fusion}
In this paper, we propose a novel approach, Recurrent-OctoMap, for long-term semantic mapping. In this approach, we maintain the occupancy and semantics within the Recurrent-OctoMap cells, and model the fusion of semantic observations as a state transition procedure. To be more specific, in addition to the cell variables introduced in Section \ref{sec:lmmm}, we further allocate the variables $state$, $prob$ for the storage of the current semantic state and the predicted probabilities for each cell. The observation of each Recurrent-OctoMap cell is the semantic feature $f_{cell}$ obtained by Eq.~\ref{eq:cell_feat}. The semantic state $state_t$ is a non-linear hidden state of observations prior to $t$, and the transition between semantic states represents a non-linear fusion of consecutive observations. 
In our approach, we model this update as a Recurrent Neural Network (here LSTM was used as we found this achieved better performance than a basic RNN \cite{Elman90findingstructure} or Gated Recurrent Unit (GRU)~\cite{DBLP:journals/corr/ChoMGBSB14} in our experiments). A recurrent loop of LSTM \cite{lstm} is:
\begin{equation}
\small
(S_{t+1}, h_{t+1}) = LSTM\big(o_t, S_{t}, h_{t}, g_i, g_f, g_o; W_{\{i,f,o,s\}} \big),
\end{equation}

\noindent where $o_t$ is the input observation at time $t$. $S_{t}, h_{t}$ and $S_{t+1}, h_{t+1}$ are the LSTM's state and output variables at time $t$ and $t+1$. $g_i, g_f, g_o$ refers to the $input$, $forget$ and $output$ gate, respectively. $W_{\{i,f,o,s\}}$ are the parameters of LSTM. 

In our approach, the $state$ of each Recurrent-OctoMap is represented as the tuple of LSTM state $S_{t}$ and the hidden state $h_{t}$. The predicted probabilities of each semantic category can be obtained by encoding $h_{t}$ with $W_e$ and a softmax layer.
\begin{equation}
\begin{split}
state_{t+1} \leftarrow update\big(f_{cell}, state_t; LSTM \big) \\
state_t = (S_{t}, h_{t}) ~~~~~~~~~~~~~~~~~~~~~~~~~~~~~~\\
prob = softmax(h_t; W_e) ~~~~~~~~~~~~~~~~~~~\\
\end{split}
\end{equation}

\noindent In the training procedure, we train the LSTM with manually annotated maps 
by minimizing the negative log likelihood. 

\subsubsection{Nap-LSTM for long-term semantic learning}
Benefiting from the $360^\circ$ horizontal FOV of the 3D Lidar sensor, each Recurrent-OctoMap cell is able to receive a much longer sequence of observations than RGB-D camera-based semantic mapping. However, in the large-scale outdoor environment, the field of view of a 3D Lidar mounted on a moving mobile robot is still constrained in some parts of the map. Underpinned by our robust long-term mapping system introduced in Section \ref{sec:long-term-slam}, the semantic observations of Recurrent-OctoMap cells can be associated in a long-term fashion. 
In our approach, a Nap-LSTM is devised for learning semantics from discretized observations along the time-axis (shown in Fig.~\ref{fig:nap-lstm}). To be more specific, when no observations are available the LSTM can ``nap'' for up to a pre-specified duration, referred to as the Maximum Napping Time Duration (MNTD). That is, the LSTM will remain in the previously observed state until either a new observation is obtained or the MNTD is reached. If the MNTD is reached, the state is reinitialized to zeros.
It is necessary to use Nap-LSTM, especially in long-term fusion, because this can prevent the state of LSTM transiting into an unknown state space. For example, if the model is trained by a pre-specified sequence length, the ordinary LSTM is able to learn the state transition within this length. However, when deployed on a longer sequence, the LSTM is likely to transit the state to an unknown state space, and consequently introduce more errors. We, therefore, introduce Nap-LSTM with MNTD parameter to manipulate the observation sequences (i.e. retain or reinitialize the state).
We investigate the sensitivity of this parameter in the experiments.

\subsubsection{Learning from long-term data}
The semantic observations of different timescales indicate different patterns of behavior. Observations of very short duration (e.g.\ a couple of seconds) show the transient dynamics, such as pedestrians/cyclists moving across a cell; observations of medium duration (e.g.\ several minutes) indicate the changes of static objects, such as parked cars perceived by the moving robot from different views; observations of long duration (e.g.\ more than a day) show the changes to the layout of the environment. An example of an associated map over three days is shown in Fig.~\ref{fig:long-term-rnn}. 

In this paper, we aim to learn the fusion of semantics from the changes along different timescales. We conduct continuous simultaneous localization, mapping and semantic understanding using two weeks of data.
As our training data is both spatially and temporally discretized, learning an individual LSTM for each cell suffers from very bad over-fitting. In order to train a generalized model from discretized training data, the weights of LSTM are shared for all the Recurrent-OctoMap cells. 
It is worth noting that the whole mapping process is trainable end-to-end as the single-frame semantic understanding is also achieved using a neural network. In this paper, we trained the mapping in separate stages due to the limitation of GPU memory. More details of the training process are provided in Section \ref{sec:experiments}.

\begin{table*}[thpb]
\centering
\caption{The quantitative result of semantic mapping for day 8, 9, 10, 11, 12, 13, 14, and the comparison with the baseline methods, including Recurrent OctoMap with Standard LSTM (Recurrent-OctoMap$^{-}$) and Bayesian update.}
\label{tab:comparison}
\begin{tabular}{|l|l|l|l|l|l|l|l|l|l|}
            \hline
             Metrics & methods & day8 & day9 & day10 & day11 & day12 & day13 & day14 & mean\\                       \hline
             \hline
            \multirow{3}{*}{Overall Acc.} & \multicolumn{1}{|l}{Recurrent-OctoMap} & \multicolumn{1}{|l}{\bf{95.6\%}}  & \multicolumn{1}{|l}{\bf{88.2\% }} & \multicolumn{1}{|l}{\bf{90.6\%}} & \multicolumn{1}{|l}{\bf{94.5\%}} & \multicolumn{1}{|l}{\bf{90.0\%}} & \multicolumn{1}{|l}{\bf{91.3\%}} & \multicolumn{1}{|l}{\bf{95.1\%}} & \multicolumn{1}{|l|}{\bf{92.2\%}} \\\cline{2-10}
                                             & \multicolumn{1}{|l}{Recurrent-OctoMap$\pmb{^{-}}$} & \multicolumn{1}{|l}{90.7\%} & \multicolumn{1}{|l}{78.3\%} & \multicolumn{1}{|l}{79.5\%} & \multicolumn{1}{|l}{84.1\%} & \multicolumn{1}{|l}{78.0\%} & \multicolumn{1}{|l}{82.4\%} & \multicolumn{1}{|l}{91.7\%}  & \multicolumn{1}{|l|}{83.5\%} \\\cline{2-10}
                                            & \multicolumn{1}{|l}{Bayesian Update} & \multicolumn{1}{|l}{80.2\%} & \multicolumn{1}{|l}{73.0\%} & \multicolumn{1}{|l}{71.2\%} & \multicolumn{1}{|l}{79.0\%} & \multicolumn{1}{|l}{70.2\%} & \multicolumn{1}{|l}{77.6\%} & \multicolumn{1}{|l}{82.2\%}  & \multicolumn{1}{|l|}{76.3\%} \\
                   \hline
                   \hline
                   
            \multirow{3}{*}{Mean Acc.} & \multicolumn{1}{|l}{Recurrent-OctoMap} & \multicolumn{1}{|l}{\bf{78.8\%} }  & \multicolumn{1}{|l}{\bf{84.9\%}} & \multicolumn{1}{|l}{\bf{76.0\%}} & \multicolumn{1}{|l}{\bf{91.4\%}} & \multicolumn{1}{|l}{\bf{80.7\%}} & \multicolumn{1}{|l}{\bf{86.3\%}} & \multicolumn{1}{|l}{\bf{77.6\%}} & \multicolumn{1}{|l|}{\bf{81.7\%}} \\\cline{2-10}
                                             & \multicolumn{1}{|l}{Recurrent-OctoMap$\pmb{^{-}}$} & \multicolumn{1}{|l}{69.5\%} & \multicolumn{1}{|l}{73.2\%} & \multicolumn{1}{|l}{63.8\%} & \multicolumn{1}{|l}{77.3\%} & \multicolumn{1}{|l}{70.4\%} & \multicolumn{1}{|l}{73.5\%} & \multicolumn{1}{|l}{70.4\%}  & \multicolumn{1}{|l|}{71.2\%} \\\cline{2-10}
                                            & \multicolumn{1}{|l}{Bayesian Update} & \multicolumn{1}{|l}{65.7\%} & \multicolumn{1}{|l}{73.7\%} & \multicolumn{1}{|l}{61.7\%} & \multicolumn{1}{|l}{82.3\%} & \multicolumn{1}{|l}{63.3\%} & \multicolumn{1}{|l}{77.3\%} & \multicolumn{1}{|l}{66.0\%}  & \multicolumn{1}{|l|}{70.0\%} \\
                   \hline
                   \hline
                   
             \multirow{3}{*}{Mean IoU} & \multicolumn{1}{|l}{Recurrent-OctoMap} & \multicolumn{1}{|l}{\bf{71.0\%}}  & \multicolumn{1}{|l}{\bf{77.4\%}} & \multicolumn{1}{|l}{\bf{65.9\%}} & \multicolumn{1}{|l}{\bf{87.0\%}} & \multicolumn{1}{|l}{\bf{68.2\%}} & \multicolumn{1}{|l}{\bf{77.6\%}} & \multicolumn{1}{|l}{\bf{64.2\%}} & \multicolumn{1}{|l|}{\bf{73.0\%}} \\\cline{2-10}
                                             & \multicolumn{1}{|l}{Recurrent-OctoMap$\pmb{^{-}}$} & \multicolumn{1}{|l}{58.2\%} & \multicolumn{1}{|l}{62.7\%} & \multicolumn{1}{|l}{50.6\%} & \multicolumn{1}{|l}{67.9\%} & \multicolumn{1}{|l}{53.4\%} & \multicolumn{1}{|l}{61.9\%} & \multicolumn{1}{|l}{54.4\%}  & \multicolumn{1}{|l|}{58.4\%} \\\cline{2-10}
                                            & \multicolumn{1}{|l}{Bayesian Update} & \multicolumn{1}{|l}{43.3\%} & \multicolumn{1}{|l}{55.0\%} & \multicolumn{1}{|l}{42.4\%} & \multicolumn{1}{|l}{62.1\%} & \multicolumn{1}{|l}{41.9\%} & \multicolumn{1}{|l}{58.3\%} & \multicolumn{1}{|l}{41.4\%}  & \multicolumn{1}{|l|}{49.2\%} \\
                   \hline          
        \end{tabular}
\end{table*}

\section{EXPERIMENTS}
\label{sec:experiments}

\subsection{ETH Parking-Lot dataset}\label{sec:dataset}

The dataset for evaluation was originally used in \cite{eth-data}. This data was collected in the parking lot of ETH for 14 consecutive days. In each day, the robot (i.e.\ LandShark system by Black-I Robotics, USA) was driven manually to explore the parking lot and back to the original position. Velodyne HDL-32E is used as the main sensor, which produces approximately 70K points per scan at a rate of 10~Hz. More details can be found in \cite{eth-data}. 

The original data only has range information (X,Y,Z) and the intensity is not available. In order to train and evaluate our proposed Recurrent-OctoMaps, we generate 14 OctoMaps (one for each day) within a global coordination system, and annotate the semantic objects manually on the OctoMaps using the L-CAS 3D Point Cloud Annotation Tool\footnote{\url{https://github.com/LCAS/cloud_annotation_tool}}. \kevinupdate{In this paper, the points from dynamic objects will remain in the map for five minutes during mapping, hence the short-term dynamics, e.g. moving cars, cyclists and pedestrians, can be mapped into these 14 OctoMaps for training and evaluation.} We annotated 4 categories of objects, i.e.\ cars, pedestrians, cyclists and the background, and the cells with overlapping dynamics are annotated as ``do not care'', which are not included in learning and validation. As we annotate the map rather than consecutive frames, the annotation effort is much less tedious. It took us approximately 30 minutes to annotate one OctoMap in this work.

\subsection{Baseline Methods}\label{sec:baseline}
Two baseline methods were implemented and used for comparison. The baseline methods share the same method in Section \ref{sec:single-scan-understanding} for the semantic understanding of a single Lidar scan, but have different mechanisms for semantic fusion.

\subsubsection{Bayesian Update}
The most widely-used fusion approach in 3D semantic mapping is the ``Bayesian update'', where the squashed probability, i.e. softmax output in Eq. \ref{eq:loss1} is used as the predictive probability for the fusion in Eq. \ref{eq:Bayesian update}. 
If no observation is available, a uniform prior distribution is assumed. This comparison shows the difference between our proposed state-based fusion and conventional end-prediction fusion.

\subsubsection{Recurrent OctoMap with Standard LSTM}
To better understand the influence of the napping mechanism, we further compared the proposed OctoMap using NapLSTM to an OctoMap with an ordinary LSTM model, i.e. in this baseline method (referred to as Recurrent-OctoMap$\pmb{^{-}}$), the napping mechanism is removed. If no observation is available, the LSTM states are initialized to zeros.

\subsection{Evaluation Metrics}
In the existing works on 3D semantic mapping~\cite{dense2014icra,oxford3d,ic2017icra,zhao2017fully}, 2D-based semantic understanding metrics are used for the evaluation, and the 3D mapping fusion is not evaluated. As the main contribution of this paper is the semantic fusion approach, we extend these performance metrics from 2D pixels to 3D voxels, including the $overall~ accuracy$ of all maps' voxels, the $mean~ accuracy$ among all semantic categories, and the $mean~ IoU$ (region intersection over union):

\begin{itemize}
\item Overall accuracy: $\sum_{i} n_{ii} / \sum_{i} {nt}_{i}$
\item Mean accuracy: $(1/n_{cl})\sum_{i} n_{ii} / {nt}_{i}$
\item Mean IoU: $ (1/n_{cl})\sum_{i} n_{ii}/({nt}_{i} + \sum_{j} n_{ji} - n_{ii}) $
\end{itemize}

\noindent where $n_{cl}$ is the number of classes, $n_{ij}$ is the number of voxels of class $i$ classified as class $j$, and $ {nt}_{i} = \sum_{j} n_{ij} $ is the total number of voxels belonging to class $i$.

\subsection{Experiments on ETH Parking-Lot Dataset}\label{sec:comparision}

\begin{figure*}[thpb]
\centering
\includegraphics[width=1.0\textwidth]{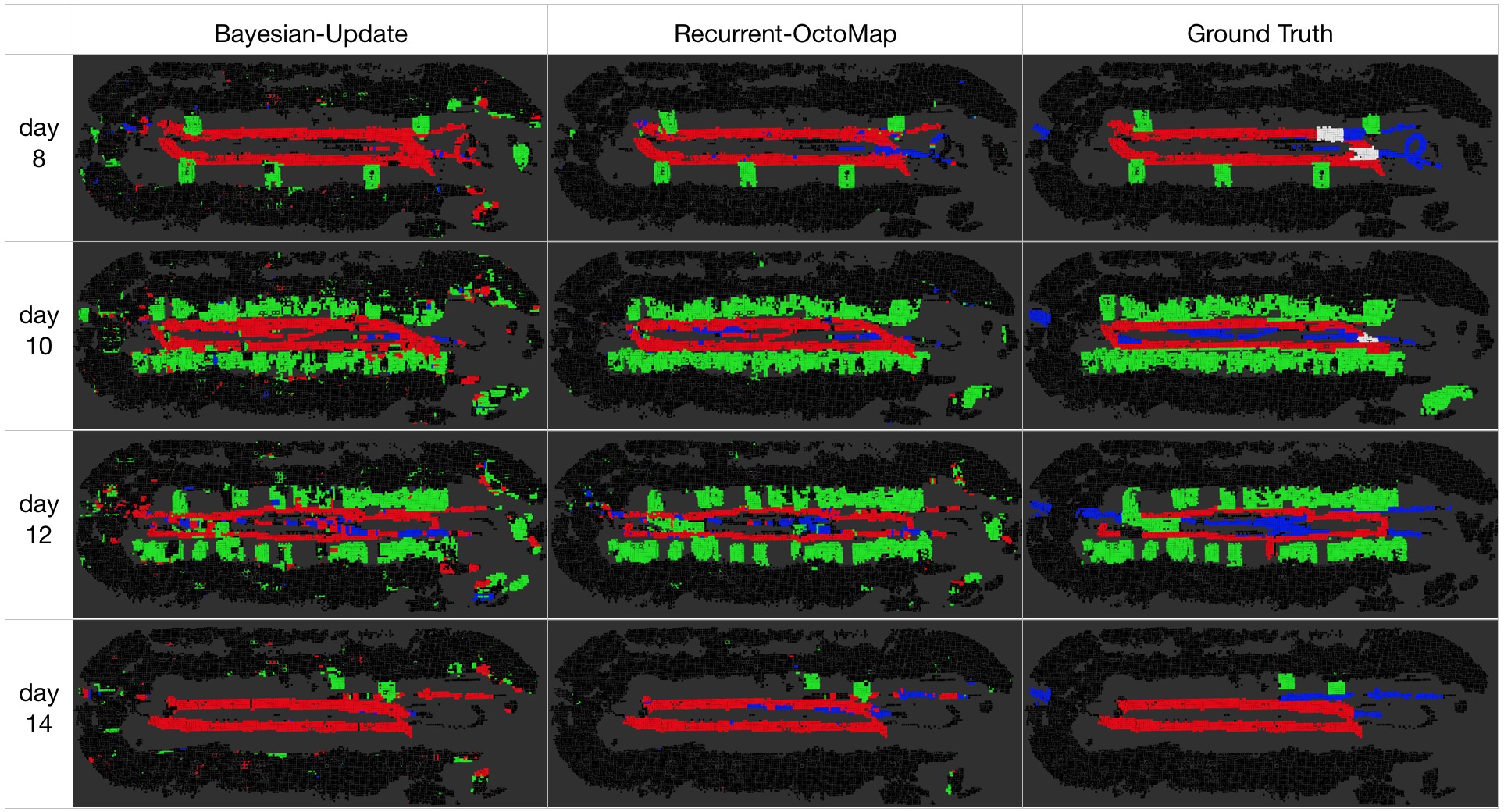}
\caption{The qualitative results of semantic mapping of Recurrent-Octomap.}
\label{fig:qualitive_results}
\end{figure*}

\subsubsection{Training}
As introduced in Section \ref{sec:dataset}, the ETH Parking Lot dataset provides 3D Lidar mapping data for 14 days. We use days 1-7 for training and days 8-14 for evaluation. The training of our proposed pipeline has two stages: training the single-scan scene understanding network and training the LSTM in Recurrent-OctoMap. To be more specific, we first pre-train the single-scan semantic understanding (i.e. detection) neural network on the KITTI 3D object challenge. In this procedure, only the bounding boxes of `cars', `pedestrians' and `cyclists' are used for training, and a total of 22524 objects and 45480 randomly selected background samples from 7481 scans are used for training. We apply random `yaw' rotations on the training scans along the z-axis as KITTI only provides front-view annotations. We train for 20 epochs with an initial learning rate of 0.005 with exponential decay of 0.95. Then we associate the scans from days 1-7 of the Parking Lot data to fine-tune the network with a smaller learning rate 0.001 for another 10 epochs. Having trained the single-scan semantic understanding network, we remove the loss function in Eq. \ref{eq:loss1} and the regularized semantic features $f_{cell}$ (obtained by Eq. \ref{eq:cell_feat}) are used as the semantic observations of Recurrent-OctoMap. A resolution of 0.4m is used for OctoMap/Recurrent-OctoMap. A two-layer Nap-LSTM cell is used as the recurrent cell of the Recurrent-OctoMap and we train the Recurrent-OctoMap as a sequence-to-sequence decoder from all semantic observations to semantic labels over the duration of 7 days. 
In other words, the MNTD (Maximum Napping Time Duration) is set as positive infinite in training. An example is given in Fig.~\ref{fig:long-term-rnn}. To be more specific, our Nap-LSTM is trained in an ``unrolled'' form with truncated backpropagation.
We train the Nap-LSTM by mini-batch with randomly sampled sequences from arbitrary start points to the end. Dynamic RNN training is used for segments less than the regularized sequence length. In this experiment, a mini-batch of 32 is used and the hidden state dimension of LSTM is 128. We train the Nap-LSTM for another 100 epochs with a learning rate 0.001 and exponential decay of 0.95. 

\subsubsection{Performance}
In the testing, we reduce the MNTD of Recurrent-OctoMap to one day in order to make the evaluation result statistically significant. 
The comparison results for the 
Recurrent-Octomap
and the two baselines 
are presented in Table~\ref{tab:comparison}. As shown in the table, Recurrent-OctoMap achieves the best performance for all the test maps, with an overall accuracy of 92.2\%, mean accuracy of 81.7\% and mean IoU of 73.0\%. It is worth noting that three categories, i.e.\ pedestrians, cars and background, are used in the evaluation for days 9, 11 and 13, as no cyclists appeared during these days. Compared to 
Recurrent-OctoMap$\pmb{^{-}}$ (without napping),
Recurrent-OctoMap experienced an improvement of approximately 9\% on overall accuracy, 10\% on mean accuracy and 15\% on mean IoU. 
Moreover, 
Recurrent-OctoMap$\pmb{^{-}}$
outperformed the Bayesian update by 7\% on overall accuracy, 1\% on mean accuracy and 10\% on mean IoU, which shows the improvement of state-based semantic fusion beyond the standard end-prediction-based fusion. Overall, our proposed Recurrent-OctoMap outperformed the baseline Bayesian Update with approximate improvements of 16\%, 12\% and 24\% on the three evaluation metrics. Qualitative results are provided in Fig.~\ref{fig:qualitive_results}. We observed that the semantic mapping error can be mostly attributed to the ill-posed detections (observations). The Bayesian Update is sensitive to incorrect observations as it simply multiplies the predictive probabilities following a first order Markov chain. Compared to the baseline approach, Recurrent-OctoMap delivers a better fusion both with and without the napping mechanism. The Recurrent-Octomap trained with sequential observations (including both good and bad observations) can learn the transitions between states and correct the predictions from the ill-posed detections. Moreover, the LSTM with the napping mechanism allows the cells to track the changing semantics in the longer-term. For example, if a cell is classified as ``car", it is likely to be ``car" again after a short duration. As a result, the predictions become more consistent, and the semantic map obtained closely matches the ground truth.

\begin{figure}[thpb]
\centering
\includegraphics[width=.5\textwidth]{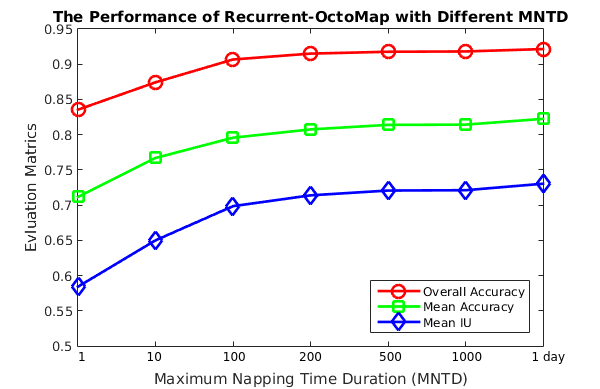}
\caption{The evaluation of Recurrent-OctoMap with different Maximum Napping Time Duration (MNTD).}
\label{fig:performance_curve}
\end{figure}

We further explore the performance of our proposed Recurrent-OctoMap with different MNTD. In this experiment, we tested the overall accuracy, mean accuracy, and mean IoU with MNTD of 1, 10, 100, 200, 500, 1000 frames and up to a day. As shown in Fig.~\ref{fig:performance_curve}, all of these three metrics experience a steep increase from 1 to 100 frames and then increase gradually within a day. These experimental results show that our proposed Recurrent-OctoMap learned the long-term state transitions from long-term mapping, and as a result, the semantic mapping performance is enhanced.
\section{CONCLUSION}\label{sec:conclusion}
In this paper, we presented a novel method Recurrent-OctoMap for state-based 3D refinement rather than prediction-based fusion in 3D semantic mapping. Compared to prediction-based fusion with Bayesian update~\cite{bayes-update}, our approach utilises latent state information modeled as a Nap-LSTM network, and is thus able to learn the semantic state transition between observations at different time-scales. We found further that the 3D Lidar-based semantic understanding and long-term localization and mapping can provide a large field of view and precise data association, which are complementary to the proposed Recurrent-OctoMap approach. In the evaluation, our proposed Recurrent-OctoMap is learned from long-term mapping data (7 days), and can maintain the semantic memory using long-term experience, which also makes the 3D semantic map more accurate. \kevinupdate{Our future work will investigate the possibility to apply the obtained recurrent-OctoMap maps to problems such as robot manipulation \cite{sun2015icra,sun2016icra}, robot localization \cite{anestis2018ral,zhao2018iros}, or human-aware navigation \cite{sun2018icra}. }

\section*{ACKNOWLEDGMENT}
We thank Prof.\ Francois Pomerleau and team for generously sharing their data. We also thank NVIDIA Corporation for donating a high-power GPU on which this work was performed. This project has received funding from the European Union's Horizon 2020 research and innovation programme under grant agreement No 732737 (ILIAD) and No 645376 (FLOBOT).

\ifCLASSOPTIONcaptionsoff
  \newpage
\fi



%
\bibliographystyle{IEEEtran}
{\bibliography{refs}}

%




\end{document}